%% file: main.tex
\documentclass[10pt,twocolumn,letterpaper]{article}

\usepackage{cvpr}
\usepackage{times}
\usepackage{epsfig}
\usepackage{graphicx}
\usepackage{amsmath}
\usepackage{amssymb}
\usepackage{subfig}
\usepackage{xcolor,soul}
\usepackage[ruled,linesnumbered,vlined]{algorithm2e}
\usepackage{fancyhdr}
\newcommand\mycommfont[1]{#1}
\SetCommentSty{\mycommfont}

\usepackage[pagebackref=true,breaklinks=true,letterpaper=true,colorlinks,bookmarks=false]{hyperref}
\newcommand{\dv}[1]{\textcolor{red}{}}
\newcommand{\dvaddr}[1]{\textcolor{blue}{}}
\newcommand{\dvtwo}[1]{\textcolor{green}{}}

\cvprfinalcopy % *** Uncomment this line for the final submission

\def\cvprPaperID{36} % *** Enter the CVPR Paper ID here

\ifcvprfinal\fi
\begin{document}

%%%%%%%%% TITLE
\title{
\textit{Now that I can see, I can improve:} \\ 
Enabling data-driven finetuning of CNNs on the edge}

\author{Aditya Rajagopal, Christos-Savvas Bouganis \\
Imperial College London\\
{\tt\small \{aditya.rajagopal14, christos-savvas.bouganis\}@imperial.ac.uk}
%\and
%Christos-Savvas Bouganis\\
%Imperial College London\\
%{\tt\small christos-savvas.bouganis@imperial.ac.uk}
}

\maketitle

\thispagestyle{fancy}
\chead{PREPRINT: Accepted at the Joint Workshop on Efficient Deep Learning in Computer Vision (EDLCV) at CVPR, 2020}
\renewcommand{\headrulewidth}{0pt}

%%%%%%%%% ABSTRACT
\begin{abstract}
    In today's world, a vast amount of data is being generated by edge devices that can be used as valuable training data to improve the performance of machine learning algorithms in terms of the achieved accuracy or to reduce the compute requirements of the model. 
    However, due to user data privacy concerns as well as storage and communication bandwidth limitations, this data cannot be moved from the device to the data centre for further improvement of the model and subsequent deployment. 
    As such there is a need for increased edge intelligence, where the deployed models can be fine-tuned on the edge, leading to improved accuracy and/or reducing the model's workload as well as its memory and power footprint.
    In the case of Convolutional Neural Networks (CNNs), both the weights of the network as well as its topology can be tuned to adapt to the data that it processes. 
    This paper provides a first step towards enabling CNN finetuning on an edge device based on structured pruning. 
    It explores the performance gains and costs of doing so and presents an extensible open-source framework that allows the deployment of such approaches on a wide range of network architectures and devices. 
    The results show that on average, data-aware pruning with retraining can provide 10.2pp increased accuracy over a wide range of subsets, networks and pruning levels with a maximum improvement of 42.0pp over pruning and retraining in a manner agnostic to the data being processed by the network. 
\end{abstract}

%%%%%%%%% BODY TEXT
\section{Introduction} \label{sec:intro}
    Modern CNN-based systems achieve unprecedented levels of accuracy in various tasks such as image recognition \cite{2020-self-training}, segmentation \cite{segmentation}, drone navigation \cite{kouris_learning_2018}, and object detection \cite{kouris_informed_2019, girshick_fast_2015} due to the vast amounts of curated \cite{barbu_objectnet_2019, imagenet} data that is used to train them.
    This training is performed on large data-centres and once deployed, the models remain static.
    However the large quantities of domain specific data that can help further improve the performance of these networks in terms of accuracy or inference time, reside on the edge. 
    This improvement can stem either from the availability of more data samples or the realisation of a different distribution of data at the deployment side. 
    Nonetheless, user data privacy concerns as well as limited storage and communication bandwidths mean that this data cannot be easily moved from the edge to these data-centres for updating the deployed model through changes to the model's architecture (i.e. topology in the case of a CNN), and its parameters.
    Consequently, there has been a push to move the required processing from data-centres to edge devices \cite{zhou_edge_2019}. 
    
    A widely adopted approach to tune the architecture of the model to the input data distribution is through pruning followed usually by a retraining stage.  
    This paper will refer to such approaches as data-aware pruning and retraining (DaPR) approaches.
    The suitability of the above approach is further supported by works such as \cite{suau_filter_distillation_2019, mittal_plasticity_2018, gao_fbs_2018, lin_runtime_neural_pruning_nodate} which have shown that the pruning levels are linked to the complexity of the data the network is processing.
    Currently the increased compute and memory capabilities of edge devices such as NVIDIA's Jetson TX2, NVIDIA's Xavier GPUs and Google's Edge TPU \cite{jouppi_tpu_nodate}, provide an opportunity to perform such tuning on edge devices in a manner that does not infringe on user data privacy and is within an acceptable time frame.
    
    With the goal to enable CNNs to improve and adapt their performance to the data they are processing on the edge, the contributions of this paper are as follows :
    \begin{enumerate}
       \item 
       A methodology based on the L1-norm of the weights that allows for on-device DaPR to be performed without user intervention. 
       In doing so, the paper explores the accuracy gains of adapting a network to the data it is processing as well as the cost of achieving these gains on an edge device.   
       The paper provides quantitative results on the possible performance gains (i.e. inference latency) and the associated costs (i.e. pruning and retraining) of after-deployment tuning on a number of state-of-the-art models targeting an actual embedded device.
       
       \item 
       An open-source framework ADaPT (\textbf{A}utomated \textbf{D}ata-aware \textbf{P}running and Re\textbf{T}raining) that allows rapid prototyping and deployment of various structured pruning techniques on a wide range of network architectures on edge devices.
       To the best of our knowledge, it is the only open-source tool that fully automates the process of identifying filters to prune, shrinking the network to obtain memory and performance gains, and performing retraining of the pruned network. In doing so, it enables direct deployment of DaPR solutions on any edge device. 
    \end{enumerate}
    
    The rest of the paper is organised as follows.
    Section \ref{sec:motivation} formally describes the field of research that this work addresses and states any assumptions made.
    Section \ref{sec:metrics} describes the metrics that are of interest when evaluating solutions within this field.
    Section \ref{sec:bkgrnd} discusses various state of the art structured pruning techniques and frameworks that allow for experimentation with pruning. 
    Section \ref{sec:solution} describes the proposed L1-norm based DaPR solution.
    Section \ref{sec:tool} describes the key features of ADaPT.
    Finally, Section \ref{sec:eval} evaluates the gains and costs of performing DaPR on an NVIDIA Jetson TX2.

\section{Motivation} \label{sec:motivation}
    Let us consider a training dataset $\mathcal{D} = \{\mathcal{I}, \mathcal{C}\}$, where $\mathcal{I}$ is the set of images in the dataset, and $\mathcal{C}$ is the set of classes represented by the images.
    Let us define a model $\mathcal{M}$ as a tuple $(\mathcal{W}, \mathcal{A})$, where $\mathcal{W}$ represents the weights of the model and $\mathcal{A}$ represents the topology of the model.
    After performing training on $\mathcal{D}$, the resulting model $\mathcal{M_\mathcal{D}} = (\mathcal{W_\mathcal{D}}, \mathcal{A_\mathcal{D}})$ is such that for class $i \in C$, $a_i$ is the accuracy that the model predicts that class with, and $c$ is the cost of the model. 
    
    Consider the case where this model $\mathcal M_\mathcal{D}$ is going to be deployed in an environment $\mathcal{D}' = \{\mathcal{I}', \mathcal{C}'\}$. 
    In most practical scenarios the classes that are expected to be seen are not completely known before deployment, but nonetheless a reasonable assumption (\textbf{Assumption 1}) that is made is that $\mathcal{D}' \subseteq \mathcal{D}$, i.e. the classes encountered upon deployment are a subset of the classes included in the training data. 
    With this in mind, the problem this work addresses is:
    \begin{quotation}
       \noindent
       Given the dataset $\mathcal{D}'$ and a provided compute and memory capacity budget, find a model $\mathcal{M_\mathcal{D}}'$, such that a) its cost $c'$ is less than the cost $c$ of the initially deployed model $\mathcal{M_\mathcal{D}}$, and b) $\frac{\sum_{j \in \mathcal{C}'} a_j'}{|\mathcal{D}'|} \geq \frac{\sum_{j \in \mathcal{C}'} a_j}{|\mathcal{D}'|}$, i.e. $\mathcal{M_\mathcal{D}}'$ performs at least as well as $\mathcal{M}_\mathcal{D}$ on $\mathcal{D}'$.
    \end{quotation}
    The transformation of a model can happen both through changes to the weights of the model and/or the topology of the model.

\section{Metrics of interest} \label{sec:metrics}
 The following metrics are adopted in order to assess the quality of various pruning strategies in this work: 
    \begin{enumerate}
        \item The achieved accuracy of the produced model.
        
        \item The cost $c$ of the produced model, which entails:
            \begin{itemize}
                \item The number of operations per second for a single inference (GOps).
                It is a widely adopted metric in literature, as it provides a platform agnostic way of comparing the inference time of various pruned networks. It should be noted that depending on the architecture of the target hardware, this is not always a good representation of the true latency of the system.
                
                \item    
                The latency of a single inference step using a specific device. 
            \end{itemize}
        
        \item The memory footprint of $\mathcal{M}_\mathcal{D}'$ as a percentage of the memory footprint of $\mathcal{M}_\mathcal{D}$. This will be referred to as the pruning level throughout the rest of this paper. 
    \end{enumerate}

\section{Background and Related Works} \label{sec:bkgrnd}
    Consider a CNN with $\mathcal{L}$ layers, where layer $l \in \mathcal{L}$ has $n_l$ convolutional filters of size ($m_l \times k_l \times k_l$) each, where $n_l$ are the number of output features, $m_l$ the number of input features, and the convolutional filter is of size $k_l \times k_l$.
    For layer $l \in \mathcal{L}$, the weight matrix $\mathbf{W}_l$ has size $n_l \times m_l \times k_l \times k_l$, and the output feature maps (OFM) are $\mathbf{X}_l$. 
    
    Unstructured pruning \cite{gale_state_2019, lin_dynamic_2020} removes individual neurons within the network and induces sparsity, requiring custom hardware optimised for sparse operations \cite{zhang_cambricon-x_2016, parashar_scnn_2017} in order to transform these changes into actual performance gains.
    Structured pruning focuses on removing entire convolutional filters and not just individual neurons.
    This allows for realising runtime gains on commercial hardware using off-the-shelf frameworks and is the approach to pruning this paper will utilise.
    
    \textbf{Structured Pruning -} The process of structured pruning generally involves pruning a pre-trained network based on a filter ranking criterion and then if possible performing finetuning of the pruned network to regain the accuracy lost due to pruning. 
    A number of works have explored various filter ranking criteria.
    \cite{li_l1_pruning_2017} uses the L1-norm of the weights to rank filters and obtains up to 64\% memory footprint and up to 38.6\% Ops reduction for negligible accuracy loss across networks on CIFAR-10. They also achieve up to 10.8\% memory and 24.2\% Ops reduction for around 1\% accuracy loss on ResNet34 on ImageNet \cite{imagenet}.
    \cite{liu_slimming_2017} use the weights learnt by batch normalisation layers to rank channels and obtain up to 29.7\% reduction in memory and 50.6\% reduction in Ops for ResNet164 on CIFAR100 and 82.5\% memory and 30.4\% Ops reduction for VGG on ImageNet for negligible accuracy loss.
    \cite{molchanov_sensitvity_importance_nodate, lecun_optimal_1990} use the concept of sensitivity which ranks filters based on an approximation of the impact on the loss that their omission has.  
    \cite{molchanov_sensitvity_importance_nodate} achieves up to 66\% memory and 2.5x latency reduction for VGG16 on ImageNet with an accuracy loss of 2.3\%.
    Sensitivity-based works require the gradient of each filter as well as the weights and are more memory intensive than the weight-based methods. 
    
    \cite{hur_entropy-based_2019} perform entropy based pruning by estimating the average amount of information from weights to output.
    They achieve up to 94\% memory reduction with negligible loss in accuracy on LeNet-5 on the MNIST\cite{lecun_mnist_1998} dataset.
    However, such information-theoretic methods are compute intensive and make them unfavourable for deployment in embedded settings.
    
    \cite{suau_filter_distillation_2019} uses a correlation based metric on the OFM ($\mathbf{X}_l$) of each layer in order to decide which filters to prune. 
    Using the feature maps makes this method input dependent, and the paper also explores the difference in filters pruned when shown only various subsets of the dataset that the original network was trained on. 
    They achieve memory reduction of 8x, 3x and 1.4x on VGG-16 for CIFAR-10, CIFAR-100, and ImageNet respectively and up to 85\% memory reduction with no accuracy loss when tuned to random subsets of CIFAR-100 with 2 classes in each subset.
    
    The works mentioned above do not dynamically choose the filters pruned based on the data that the network is processing.
    There is a line of works that reduce the required operations during inference by skipping convolution operations depending on the input image by introducing conditional execution. 
    However, these works do not reduce the memory requirements of the model.
    \cite{gao_fbs_2018, hua_channel_gated_2019, lin_runtime_neural_pruning_nodate} all pre-train classifiers that, at run-time, can identify which filters are important for the current input image. 
    For VGG-16, \cite{gao_fbs_2018} achieves a 1.98x Ops reduction for a 2\% decrease in accuracy on ImageNet and show that they outperform both \cite{hua_channel_gated_2019} and \cite{lin_runtime_neural_pruning_nodate} on the same metric. 
    \cite{zhang_scan_2019} splits the network into multiple sections and learns classifiers that allow for early exit through the network depending on the input image processed.
    They achieve on average 2.17x reduction in Ops across networks on CIFAR-100 for no accuracy loss, and 1.99x reduction in Ops on ImageNet also for no accuracy loss. 
    
    \textbf{Frameworks -} The various pruning techniques discussed above each have a unique set of hyperparameters that relate to filter ranking metrics as well as the manner in which the models are re-trained. 
    For instance, \cite{suau_filter_distillation_2019} sequentially prunes and retrains on a per layer basis, while works such as \cite{zhang_scan_2019} have to add many auxiliary layers on top of the chosen architecture in order to create and train their early exit classifiers. 
    Distiller \cite{zmora_distiller_2019} and Mayo \cite{zhao_mayo:_2018} are two state-of-the-art open-source frameworks that allow for experimentation with such pruning techniques.  
    Mayo focuses on automating search for hyperpameters related to pruning, while Distiller focuses on implementing a wide variety of pruning techniques discussed above.  
    Distiller provides a functionality called "Thinning" for ResNet models only, but does not allow for easy application to other networks as the functionality is tailored to the ResNet architecture. 
    Moreover, both frameworks do not automatically shrink the size of the model after pruning, but instead mask the weights of the model in order to allow for experimentation with the pruned model.
    Consequently, they can only be used to assess the impact of pruning on the accuracy and not on run-time. 
    In contrast, ADaPT addresses both these issues by shrinking the model for a wide variety of architectures to help realise run-time gains, as well as providing an extensible codebase to apply model shrinking to any new architecture. 

\section{On-device DaPR Methodology} \label{sec:solution}
    This section presents an approach to obtain a model $\mathcal{M_D}'$ from $\mathcal{M_D}$ as described in Section \ref{sec:motivation}. It is assumed that the inputs collected upon deployment of the system have been correctly classified (\textbf{Assumption 2}). The assumption enables training to be performed on the edge without uncertainty of the class-labels, and thus focusing solely on gains that can be made by adapting the network to the data it processes upon deployment.
    
    \subsection{Search Methodology} \label{sec:pp-search}
        \input{algorithms/dapr}
        Adapting the network architecture to the data it is processing involves searching for a model $\mathcal{M}_\mathcal{D}'$ that performs at least as well as $\mathcal{M}_\mathcal{D}$ on $\mathcal{D}'$, but with a reduced cost.
        The proposed approach is shown in Algorithm \ref{alg:dapr} and performs a binary search over a range of predefined pruning levels. 
        If progressively larger pruning levels are searched, the time to perform this search and the memory footprint of searched models decreases as progressively smaller models are used. 
        The algorithm converges when the pruning level to be searched does not change over iterations of the binary search.

\section{ADaPT} \label{sec:tool}
    In order to support the quick development and easy deployment of DaPR methodologies on edge devices, the ADaPT framework has been developed. In its current state it implements, deploys and evaluates the methodology described in Section \ref{sec:solution}, but its extensible nature allow the deployment of other DaPR methodologies overcoming the limitations of the existing tools described in Section \ref{sec:bkgrnd}. 
    A high-level description of its important features are presented below, where further details on how to use it can be found on the github \footnote{\href{https://github.com/adityarajagopal/pytorch\_training.git}{https://github.com/adityarajagopal/pytorch\_training.git} \label{link:github}}.
    ADaPT's functionality can be split into 4 stages; 1) pruning dependency calculation, 2) pruning, 3) model writing, and 4) weight transfer.

    \subsection{Pruning Dependency Calculation (PDC)}
        \input{figures/special_blocks}

        Modern CNN networks are usually constructed through a set of structural modules connected in a specific way. Most of the networks are based on the modules found in AlexNet \cite{krizhevsky_alexnet_2017}, ResNet20 \cite{he_resnet_2015}, MobileNetV2 \cite{sandler_mobilenetv2_2019}, and SqueezeNet \cite{iandola_squeezenet_2016}. These four networks incorporate amongst them the most commonly used structural modules - \textit{Sequential connectivity} in AlexNet, \textit{Residuals} in ResNet20 and MobileNetV2, \textit{MBConv modules} in MobileNetV2 and \textit{Fire modules} in SqueezeNet. For instance, EfficientNet \cite{tan_efficientnet_2019} uses the \textit{MBConv} as its primary convolutional module. 
        
        Moreover, certain connectivity patterns result in pruning dependencies that necessitate the pruning of identical filters across dependent layers.
        \dvtwo{I would probs put across layers to indicate that all layers get affected}
        The PDC automates the process of recognising such dependencies within a network.
        
        Networks with only \textit{Sequential connectivity} patterns such as AlexNet and VGG \cite{simonyan_vgg_2015} do not have any pruning dependencies. 
        The same applies to \textit{Fire modules} that make up SqueezeNet.
        
        ResNet variants are made of residual blocks as shown in Fig.\ref{fig:res_block}. 
        Due to the residual connection and summation following it, the same filters need to be pruned in the last convolution ($conv_{final}$) of each residual block. 
        Works such as \cite{lin_resnet_pruning_ignore_2019} choose to not prune the last convolutional layer while others such as \cite{li_l1_pruning_2017} choose to enforce this dependency.  
        Another consideration with ResNet architectures is that the network is split into groups of residuals where with every new group, a downsampling 1x1 convolution ($conv_{down}$) is added to the residual connection which ensures the number of channels output from the connection match those output from $conv_{final}$. 
        In these cases, the pruning dependency exists between $conv_{final}$ and $conv_{down}$. 
        The PDC enforces either a dependency across all $conv_{final}$ layers within a group of residual blocks or prunes $conv_{down}$ in line with its corresponding $conv_{final}$.
        
        MobileNetV2 is made of MbConv blocks which contain depth-wise separable convolutions as shown in Fig.\ref{fig:dw_conv_block} where the blue blocks are input feature maps (IFM) and the red blocks are convolutional filters. 
        The 3x3 convolution in the figure is the depth-wise convolution ($conv_{dw}$) and is different from regular convolutions in that each filter only acts on one of the IFM, i.e. $n_l$ = 1.
        This means that the number of filters in $conv_{dw}$ must always match the number of IFM to that layer, and consequently the same filters need to be pruned in $conv_{dw}$ and the layer(s) feeding it.  
        This dependency is also enforced by the PDC when MBConv modules are present in the model.
        
        The PDC takes a model description in PyTorch that the user annotates with Python decorators to identify classes that correspond to various structural modules and the names of the convolutions within them.
        Based on this information, the PDC automatically calculates all the dependent layers and communicates this information to the pruning stage so layers can be pruned in dependent groups if necessary.
        This automation makes ADaPT very easy to use as tools such as Distiller \cite{zmora_distiller_2019} require the user to manually identify each dependent convolution in the entire network, which for larger networks can be very tedious to list. 
        
    \subsection{Pruning}
        Pruning the network once the dependencies have been identified is performed by ranking all the filters based on a customised metric, and then removing filters one-by-one until the desired percentage of memory has been achieved. 
        The user could chose to rank filter globally or on a per layer basis.
        The block utilises the PDC information about dependencies between layers that need to be considered when pruning.
        Removing a filter from one of the dependent layers removes one from all of the layers in that dependency chain. 
        Furthermore, the effect of removing a filter is propagated through the network as each filter corresponds to an IFM for the next layer. 
        This stage computes the filters per layer that need to be pruned and passes this information to the Model Writing stage. 
    
    \subsection{Model Writing}
        In order to effectively evaluate the solution described in Section \ref{sec:solution}, a necessary ability of the tool is to produce a new model that has a reduced runtime and memory footprint, so when possible the $n_r$ epochs of retraining of a pruned model can be accelerated.
        The Model Writing stage enables this by creating a new shrunk PyTorch model description having removed the filters that were pruned. 
        
        The Model Writer takes the channels to be pruned as input, and provides the user with a description of a pruned network which can be used in any PyTorch code-base. 
        This decouples model pruning from the model writing and allows for easy access to pruned models.  
    
    \subsection{Weight Transfer}

\input{tables/struc_modules}

        The final stage transfers the relevant weights from $\mathcal{M}_\mathcal{D}$ to $\mathcal{M}_\mathcal{D}'$.
        This is necessary to minimise the drop in accuracy that is seen once the network is pruned and thus minimise the number of epochs $n_r$ in the retraining stage that are necessary to reach the target performance.
        It takes the pruned model description as input and returns a PyTorch model that is ready for deployment with the transferred weights.
    \\\\
    It is important to note that the parts discussed in this section are linked to structural modules and not networks. This makes the tool more extendable as any network that contains the above structural modules can be readily pruned with the current version of the tool. 
    Furthermore, it is built for easy customisation of all the functions discussed above, thus allowing for new architectures and different pruning techniques to be experimented on with ease. 
    A summary of the supported structural modules, as well as the networks that contain these modules and hence supported by ADaPT are presented in Table \ref{tab:struc_modules}. 

\section{Evaluation} \label{sec:eval}
    A key contribution of this work is the analysis of the cost reduction on model deployment obtained by performing on-device DaPR as well as the associated costs of performing such domain adaptation on an edge device.
    The methodology being evaluated is that presented in Section \ref{sec:solution}. 
    The results reported in this section are averages and standard deviations across 5 independent runs of each experiment.
    
    \subsection{Setup and Hyperparameters} \label{sec:train-setup} 
        The chosen edge device was the NVIDIA Jetson TX2. 
        Through cross-validation, the values chosen for $n_f$ and $n_r$ were 5 and 25 respectively to ensure that the accuracy plateaus before retraining ends.  
        The range of pruning levels searched were from $p_l = 5\%$ to $p_u = 95\%$ in increments of $p_i = 5\%$.
        The first pruning level searched $p_0 = 50\%$.
    
        \input{tables/subsets}
        In order to explore the problem setup discussed in Section \ref{sec:motivation}, it is necessary to create subsets $\mathcal{D'} \subseteq \mathcal{D}$. 
        In this case, $\mathcal{D}$ was the entire CIFAR-100 dataset, and five different subsets $\mathcal{D'}$ were tested. 
        CIFAR-100 is categorised into 20 "coarse-classes" which contain 5 "fine-classes" each. 
        The subsets created and the classes within them are described in Table \ref{tab:subsets}.
        The first four subsets were hand selected to have coherent semantic meaning and across the four of them span the entire CIFAR-100 dataset. 
        The fifth subset was randomly generated.
        The networks tested on were AlexNet, ResNet20, MobiletNetV2, and SqueezeNet for all the subsets listed in Table \ref{tab:subsets}.
        
        The chosen metric to rank filters was the L1-norm as it has been established to show competitive performance even against data-aware metrics \cite{mittal_plasticity_2018} despite being relatively computationally inexpensive. 
        
        The learning rate schedule was set to start at the final learning rate employed when $\mathcal{M}_\mathcal{D}$ was trained on $\mathcal{D}$. 
        After pruning the learning rate was increased to the second highest learning rate that was employed when $\mathcal{M}_\mathcal{D}$ was trained on $\mathcal{D}$. 
        Following this, the learning rate was decayed at epochs 15 and 25 by the same gamma that was used when $\mathcal{M}_\mathcal{D}$ was trained on $\mathcal{D}$. 
        % [\textbf{TODO} add quick summary of nf lr, nr lr0 and gamma for each network]
        
        The batch size used was 128, and the training dataset was split into a training and validation set in the 80:20 ratio. 
        The accuracy values reported in this section are the test set accuracy corresponding to the model with the best validation accuracy. 
        Standard data augmentation techniques for CIFAR-100 were used for the training set such as random crop, rotation and flip.
    
    \subsection{Performance Gains and Cost Analysis} \label{sec:perf_gains_cost}
        \input{figures/testacc_vs_inference_time}
        Fig.\ref{fig:inftime} shows the trade-off between the achieved test accuracy and inference time for all pruning levels between 5\% and 95\% on various subsets and networks.
        The red dot in each of the figures (\textbf{Unpruned}) displays the performance of $\mathcal{M}_\mathcal{D}$ on $\mathcal{D}'$. 
        The orange dots in the figures (\textbf{Subset Agnostic Pruning}) display the performance of a model that was pruned without any finetuning on $\mathcal{D}'$ and retrained on the entirety of $\mathcal{D}$ for $n_f + n_r$ epochs.
        These sets of point serve as baselines to compare the proposed DaPR methodology as they do not finetune the network being deployed to the domain ($\mathcal{D}'$) in any way, i.e. they are subset agnostic methods. 
        The green points (\textbf{Subset Aware Pruning}) display the performance of a model that was finetuned for $n_f$ epochs on the subset $\mathcal{D}'$, then pruned and subsequently retrained on $\mathcal{D}'$ for $n_r$ epochs.
        Instead of performing a binary search as proposed in Section \ref{sec:solution}, Fig.\ref{fig:inftime} acts as an "oracle" that for each pruning level compares the performance of the data-aware method (green points) with the data agnostic methods (orange and red points). 
        
        Figs.\ref{fig:alexnet_ebar}-\ref{fig:squeezenet_ebar} use errorbars to display the mean and standard deviation of the test accuracy across all 5 subsets for a given pruning level and network. 
        For many cases, the worst performing subset aware strategy performs better than the best performing subset agnostic strategy with an average improvement in test accuracy of 10.2pp and maximum improvement of 42.0pp over all pruning levels, networks and datasets.
        Furthermore, this improvement in test accuracy tends to increase as more of the network is pruned (lower inference time) thus further motivating the need to perform data-aware pruning and retraining as better accuracy can be achieved for smaller models.
        
        Figs.\ref{fig:alexnet_aquatic}-\ref{fig:squeezenet_indoors} show the same trade-off but for specific combinations of network and subset. 
        From left to right, the size of the subset increases and intuitively the gap between subset-aware and subset-agnostic pruning and retraining decreases as the subset $\mathcal{D'}$ converges towards $\mathcal{D}$.
        Nonetheless for all pruning levels, the performance of the subset-aware strategy outperforms subset-agnostic pruning and finetuning.
    
        \input{figures/search_inf_tradeoff}
        The "oracle" results discussed above show the gains that can be obtained for a wide range of pruning levels, however only some of the models perform better than the baseline performance of $\mathcal{M}_\mathcal{D}$ on $\mathcal{D}'$ (red points). 
        The methodology proposed in Section \ref{sec:solution} efficiently searches for the low inference time models that can perform better than $\mathcal{M}_\mathcal{D}$ on $\mathcal{D}'$ on an edge device.  
        Figs.\ref{fig:alexnet_random_search}-\ref{fig:squeezenet_natural_search} each form a pareto-frontier describing the trade-off between searching for a smaller model and the inference time of this model upon deployment. 
        The labels next to the points show the relative improvement in GOps compared to the unpruned model (-x times) and the pruning level of the model (memory footprint reduction).
        
        Across all subsets of the CIFAR-100 dataset, performing the search described in Section \ref{sec:solution} results in sub-minute search times per minibatch for up to 2.22x improvement in inference times, 4.18x improvement in GOps and 90\% memory footprint reduction. 
        Furthermore, the memory utilisation of the GPU does not exceed ~2GB which lies far below the maximum memory availability of the TX2 of 8GB.
        The number of minibatches searched for will depend on both the amount of data available and the time budget allocated during deployment to perform finetuning. 
        However, the results presented here show that such DaPR methodologies can be performed on edge devices within a reasonable time budget. 
        
        It should be noted that as the budget allocated for performing the pruning in this work is much shorter than the budget required by the current state-of-the-art pruning methods described in Section \ref{sec:bkgrnd}, and as such a direct comparison to those works is not meaningful.
        The existing methods perform pruning and retraining before deployment, and do not actively adapt the network once deployed.
        % To the best of our knowledge this is the only work that prunes networks on such a limited memory and resource budget since all processing occurs on an NVIDIA Jetson TX2. 
        Furthermore, none of these works except \cite{suau_cifar_subset_filter_distill_2019} report results on subsets of CIFAR-100. 
        \cite{suau_cifar_subset_filter_distill_2019} however does not provide details of the classes present in each subset, making a direct comparison infeasible.
    
    \subsection{Filter Selection} \label{sec:filter-selection} 
        \input{figures/channel_diff}
        An investigation was carried out on the relationship between the type of the structural block and the impact of finetuning to its parameters.
        Fig.\ref{fig:all_nets_aquatic_cdiff} shows the percentage difference between the filters that were selected to prune if pruning were to take place at epoch 0 and at epoch $n_f$ after $n_f$ epochs of finetuning on $\mathcal{D}'$.
        In this case $\mathcal{D}'$ is the \textit{Aquatic} subset.  
        The results show that weights in ResNet20 and MobileNetV2 are highly sensitive to finetuning on a subset, but AlexNet and SqueezeNet show negligible change in the filters selected to prune before and after the finetune stage.
        
        To analyse if these changes in selected filters are due to the finetuning process or random variation of the weights during training, ResNet20 and MobileNetV2 were further explored.
        The blue bars in Figs.\ref{fig:resnet20_aquatic_cdiff}-\ref{fig:mobilenet_random_cdiff} show the difference in channels pruned at epoch 0 versus epoch $n_f$. 
        The $n_f$ epochs of finetuning was performed 5 times per network and subset from the same starting point, thus resulting in 5 models at epoch $n_f$. 
        The orange bars show the results for the average difference in channels pruned across all $5 \choose 2$ combinations of models at epoch $n_f$.
        Fig.\ref{fig:resnet20_aquatic_cdiff} shows that with ResNet architectures, there is a large random variation due to training (high percentage orange bars), and comparable percentages between the orange and blue bars suggest that the effect of the finetuning process in tuning the topology of the network to the data processed is limited. 
        However for the MobiletNetV2 architecture, Figs.\ref{fig:mobilenet_indoor_cdiff} - \ref{fig:mobilenet_random_cdiff} show that the finetuning process effectively tunes the topology to the data processed. 
        
        These results suggest that Residual structural blocks (common feature between ResNet and MobileNetV2) make the weights sensitive to finetuning, but the depth-wise convolution (unique to MobileNetV2) allows for data-aware discrimination between filters based on just their L1-norm.
        However, further experiments need to be conducted to generalise such behaviour but the results also suggest that the metric that needs to be used for data dependent tuning of architectures may vary depending on the structural blocks present.

\section{Conclusion} \label{sec:conc}
    Acknowledging the shift in paradigm from server based compute to increased edge processing, this work provides a solution that performs on-device DaPR, an analysis of the accuracy gains achievable by performing such data-aware retraining, the costs of performing this process on an embedded device, and a tool that alows for rapid deployment and research of on-device DaPR methodologies.  
    The results show that the gains in accuracy obtained by retraining to the subset are significant and on average 10.2pp across a wide range of subsets, networks and pruning levels.
    In terms of costs, the search for a pruned model that achieves a given target accuracy can be performed on an NVIDIA Jetson TX2 with sub-minute search times per minibatch for up to a 2.22x improvement in inference latency and 90\% reduction in memory footprint.
    Furthermore, analysis of the selected filters show that for the MobileNetV2 architecture, the L1-norm is an effective yet computationally inexpensive metric to tune a network's topology to the data being processed and suggests that different architectures may require different metrics to make pruning of the network dependent on the data it is processing.
    
    Additionally, the extensible and customisable tool (ADaPT) developed to perform this on-device pruning and retraining allows users to prune a wide range of CNN architectures and realise the memory and runtime gains immediately on any platform of their choice in a fully automated manner. 
    To the best of our knowledge, this is the only open-source tool that that allows the user to automatically shrink (through pruning) and deploy a network for such a wide variety of networks as well as target hardware.
    There are commercial tools \cite{kaiyuan_guo_deephi_2016} that can perform this function, however tend to focus deployment on specific target hardware architectures such as FPGAs.
    These combination of features allow for rapid deployment of pruned models on any device without user intervention and thus makes it a unique open-source tool for pruning research. 

{\small
\bibliographystyle{ieee_fullname}
\bibliography{egbib}
}

\end{document}

%% file: algorithms/dapr.tex
\begin{algorithm}
   \label{alg:dapr}
   \caption{Proposed DaPR Methodology}
   \SetAlgoLined 
   \SetKwInOut{KwIn}{Inputs}
   \KwIn{Initial model $\mathcal{M}_\mathcal{D}$, Subset $\mathcal{D}'$, First pruning level to search $p_0$, Set of pruning levels to search $\mathcal{P} = \{ ip_i | i \in [\frac{p_l}{p_i}, \frac{p_u}{p_i}] \}$, Target Accuracy $a_{target}$}
   \KwOut{$\mathcal{M}_\mathcal{D}'$ with test accuracy $a_{max}$ and pruning level $p_0 \in \mathcal{P}$ that is the highest pruning level s.t. $a_{max} \geq a_{target}$}
   
    Finetune $\mathcal{M}_\mathcal{D}$ for $n_f$ epochs on $\mathcal{D}'$ to obtain $\mathcal{M}_\mathcal{D}^f$ 
    
    \While{$p_0$ changes across iterations}{
        Prune $\mathcal{M}_\mathcal{D}^f$ to pruning level $p_0$ according to chosen pruning strategy\\
        Retrain the pruned model for $n_r$ epochs on $\mathcal{D}'$ and record model $\mathcal{M}_\mathcal{D}'$ and corresponding test accuracy $a_{max}$ for the model with highest validation accuracy\\
        \eIf{$a_{max} < a_{target}$}{
            $p_u = p_0$\\
            \tcp{Reduce pruning level to the nearest multiple of $p_i$ to the midpoint between $p_l$ and $p_0$}
            \tcp{Reduce pruning level to the nearest multiple of $p_i$ to the midpoint between $p_l$ and $p_0$}
            $p_0 = p_i \times \lceil \frac{ \lfloor \frac{p_l + p_0}{2} \rfloor}{p_i} \rceil$
        }{
            $p_l = p_0$\\ 
            \tcp{Increase pruning level to the nearest multiple of $p_i$ to the midpoint between $p_u$ and $p_0$}
            $p_0 = p_i \times \lceil \frac{ \lfloor \frac{p_u + p_0}{2} \rfloor }{p_i} \rceil$
        } 
    } 
\end{algorithm}

%% file: figures/special_blocks.tex
%\begin{figure*}[t!]
%    \centering
%    \begin{tabular}{cc}
%        \subfloat[Residual Blocks \cite{resnet_he_deep_2015}]{\includegraphics[width = 0.45\textwidth]{figures/architectures/res_blocks.png} \label{fig:res_block}} & 
%        \subfloat[Depthwise Separable Unit \cite{sandler_mobilenetv2_2019}]{\includegraphics[width = 0.25\textwidth]{figures/architectures/dw_sep_block.png} \label{fig:dw_conv_block}} 
%    \end{tabular}
%    \caption{Structural blocks that require consideration of dependencies when pruning}
%    \label{fig:special_blocks}
%\end{figure*}

\begin{figure}[h]
    \centering
    \begin{tabular}{c}
        \subfloat[Residual Blocks \cite{resnet_he_deep_2015}]{\includegraphics[width = 0.45\textwidth]{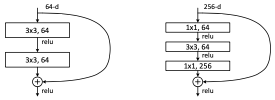} \label{fig:res_block}} \\ 
        \subfloat[Depthwise Separable Unit \cite{sandler_mobilenetv2_2019}]{\includegraphics[width = 0.25\textwidth]{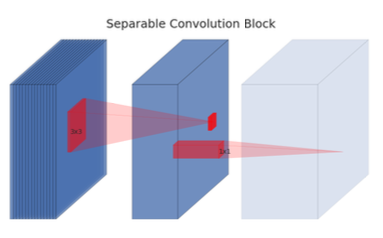} \label{fig:dw_conv_block}} 
    \end{tabular}
    \caption{Structural blocks that require consideration of dependencies when pruning}
    \label{fig:special_blocks}
\end{figure}

%% file: tables/struc_modules.tex
\begin{table}[t]
\centering
\begin{tabular}{|l|l|}
\hline
\multicolumn{1}{|c|}{\textbf{Structural Module}} & \multicolumn{1}{c|}{\textbf{Networks with Module}} \\ \hline
Sequential & AlexNet\cite{krizhevsky_alexnet_2017}, VGG\cite{simonyan_vgg_2015} \\ \hline
Residuals & \begin{tabular}[c]{@{}l@{}}ResNet\cite{resnet_he_deep_2015}, MobileNetV2\cite{sandler_mobilenetv2_2019},\\ ResNeXt\cite{xie_resnext_2017}, DenseNet\cite{huang_densenet_2018}\end{tabular} \\ \hline
Depth-wise Separable & \begin{tabular}[c]{@{}l@{}}MobileNetV2\cite{sandler_mobilenetv2_2019},\\ Xception\cite{chollet_xception_2017}, EfficientNet\cite{tan_efficientnet_2019}\end{tabular} \\ \hline
Fire Module & SqueezeNet\cite{iandola_squeezenet_2016}, GoogLeNet\cite{szegedy_googlenet_2014} \\ \hline
\end{tabular}
\caption{A summary of the structural modules implemented along with networks that contain each module.}
\label{tab:struc_modules}
\end{table}

%% file: tables/subsets.tex
\begin{table*}[t]
\centering
\begin{tabular}{|c|l|c|}
\hline
\textbf{\begin{tabular}[c]{@{}c@{}}Subset\\ Name\end{tabular}} &
  \multicolumn{1}{c|}{\textbf{\begin{tabular}[c]{@{}c@{}}Coarse Classes\end{tabular}}} &
  \textbf{\begin{tabular}[c]{@{}c@{}}\# Fine\\ Classes\end{tabular}} \\ \hline
Aquatic &
  aquatic\_mammals, fish &
  10 \\ \hline
Indoor &
  food\_containers, household\_electrical\_devices, household\_furniture &
  15 \\ \hline
Outdoor &
  \begin{tabular}[c]{@{}l@{}}large\_man-made\_outdoor\_things, large\_natural\_outdoor\_scenes, vehicles\_1, vehicles\_2, \\ trees, small\_mammals, people\end{tabular} &
  35 \\ \hline
Natural &
  \begin{tabular}[c]{@{}l@{}}flowers, fruit\_and\_vegetables, insects, large\_omnivores\_and\_herbivores, medium\_mammals, \\ non-insect\_invertebrates, small\_mammals, reptiles\end{tabular} &
  40 \\ \hline
Random &
  \begin{tabular}[c]{@{}l@{}}aquatic\_mammals, fish, flowers, fruit\_and\_vegetables, household\_furniture, \\ large\_man-made\_outdoor\_things, large\_omnivores\_and\_herbivores, medium\_mammals, \\ non-insect\_invertebrates, people, reptiles, trees, vehicles\_2\end{tabular} &
  65 \\ \hline
\end{tabular}
\caption{Subsets tested along with the coarse classes that were included in the subset and the number of fine classes per subset}
\label{tab:subsets}
\end{table*}

%% file: figures/testacc_vs_inference_time.tex
\begin{figure*}[t!]
    \centering
    \begin{tabular}{cccc}
        \subfloat[AlexNet]{\includegraphics[width = 0.25\textwidth]{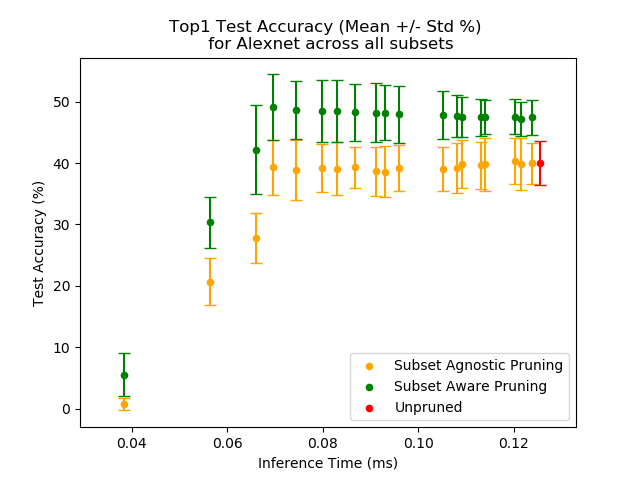} \label{fig:alexnet_ebar}}
        \subfloat[ResNet20]{\includegraphics[width = 0.25\textwidth]{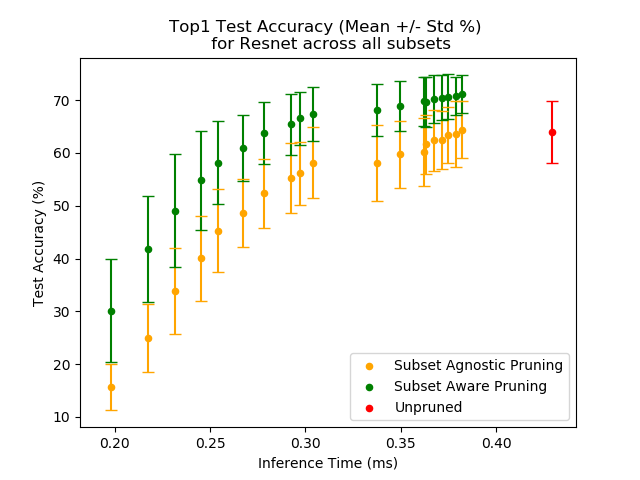} \label{fig:resnet_ebar}} 
        \subfloat[MobileNetV2]{\includegraphics[width = 0.25\textwidth]{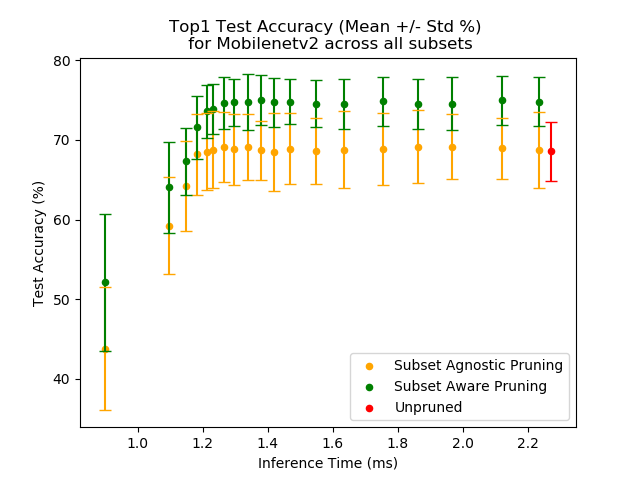} \label{fig:mobilenetv2_ebar}} 
        \subfloat[SqueezeNet]{\includegraphics[width = 0.25\textwidth]{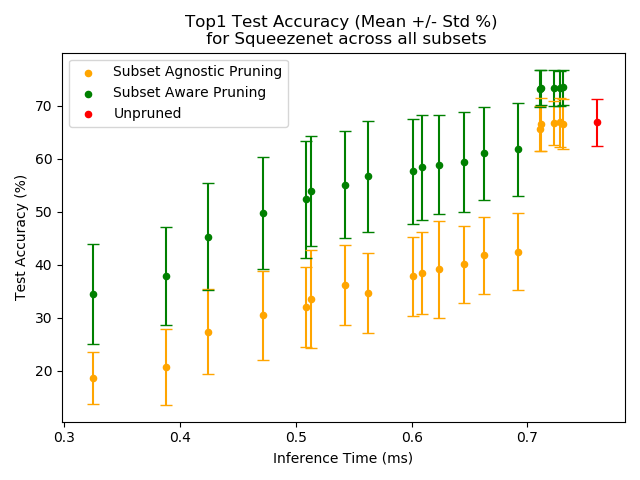} \label{fig:squeezenet_ebar}} \\
        
        \subfloat[AlexNet - Aquatic]{\includegraphics[width = 0.25\textwidth]{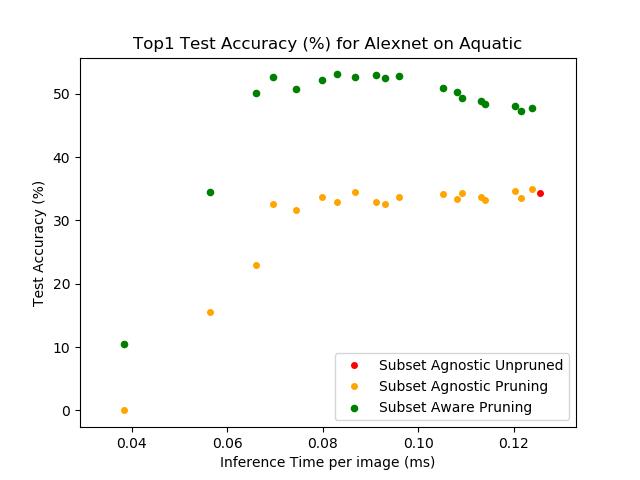} \label{fig:alexnet_aquatic}}
        \subfloat[SqueezeNet - Indoor]{\includegraphics[width = 0.25\textwidth]{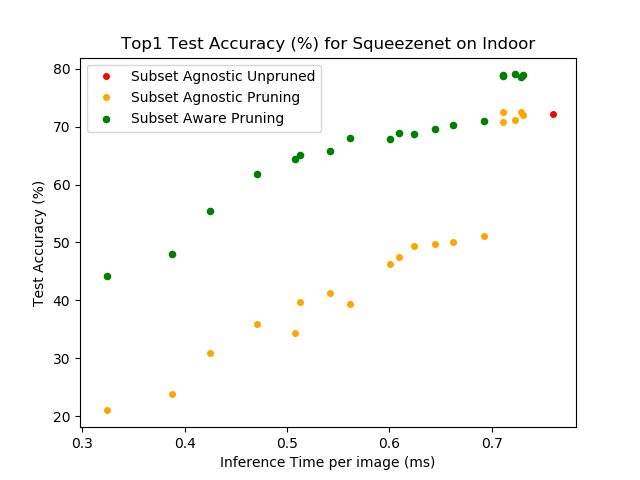} \label{fig:squeezenet_indoors}} 
        \subfloat[MobileNetV2 - Outdoor]{\includegraphics[width = 0.25\textwidth]{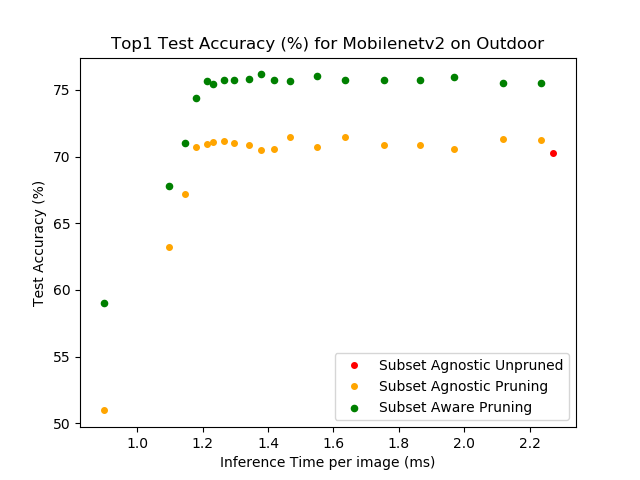} \label{fig:mobilenetv2_subset1}} 
        \subfloat[ResNet20 - Random]{\includegraphics[width = 0.25\textwidth]{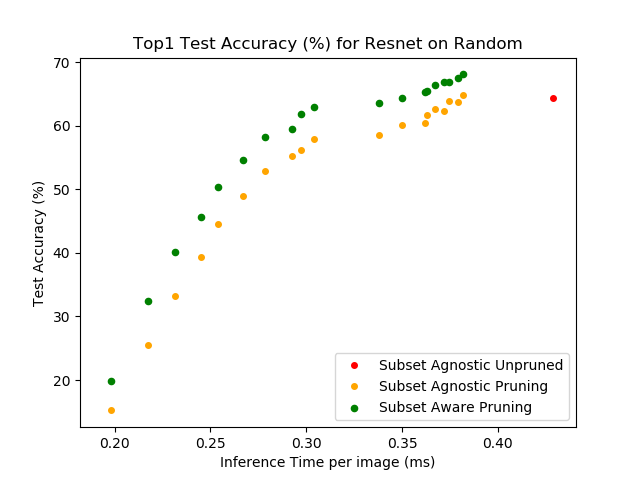} \label{fig:resnet_random}} 
    \end{tabular}
    \caption{Trade-off of Test Accuracy vs Inference Time for various levels of pruning. The error bars show the mean and standard deviation of test accuracy obtained per level of pruning across all the 5 subsets.}
    \label{fig:inftime}
\end{figure*}

%% file: figures/search_inf_tradeoff.tex
\begin{figure*}[t!]
    \centering
    \begin{tabular}{cccc}
        \subfloat[AlexNet - Random]{\includegraphics[width = 0.25\textwidth]{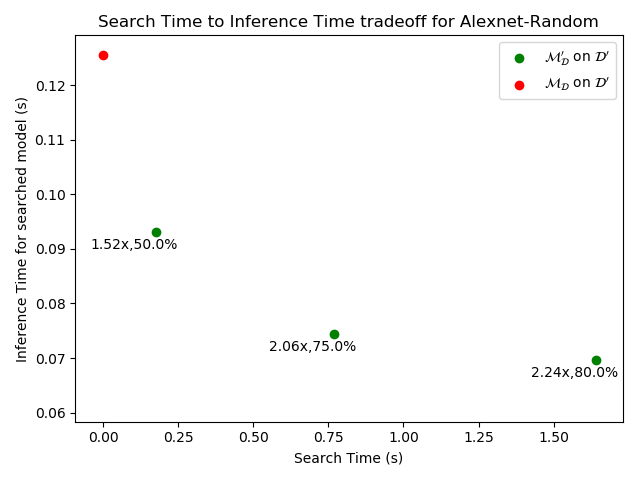} \label{fig:alexnet_random_search}}
        \subfloat[ResNet20 - Aquatic]{\includegraphics[width = 0.25\textwidth]{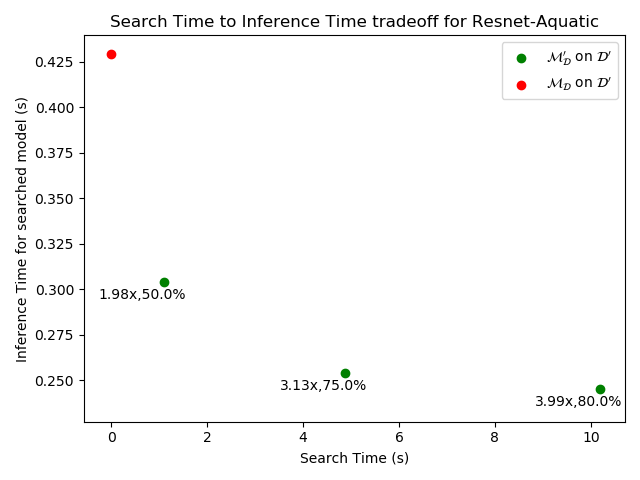} \label{fig:resnet_aquatic_search}} 
        \subfloat[MobileNetV2 - Outdoor]{\includegraphics[width = 0.25\textwidth]{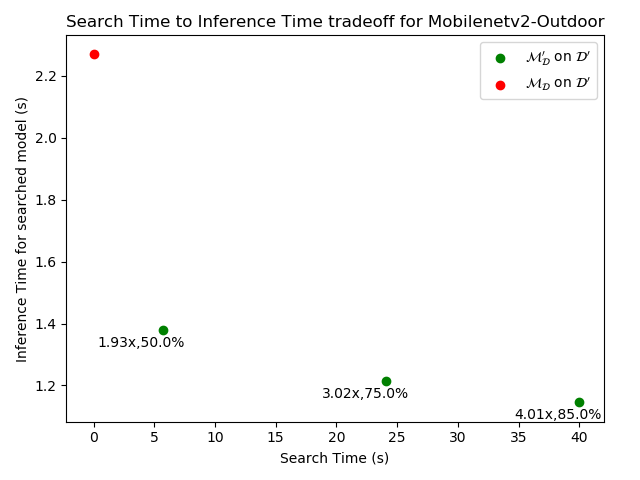} \label{fig:mobilenetv2_outdoor_search}} 
        \subfloat[SqueezeNet - Natural]{\includegraphics[width = 0.25\textwidth]{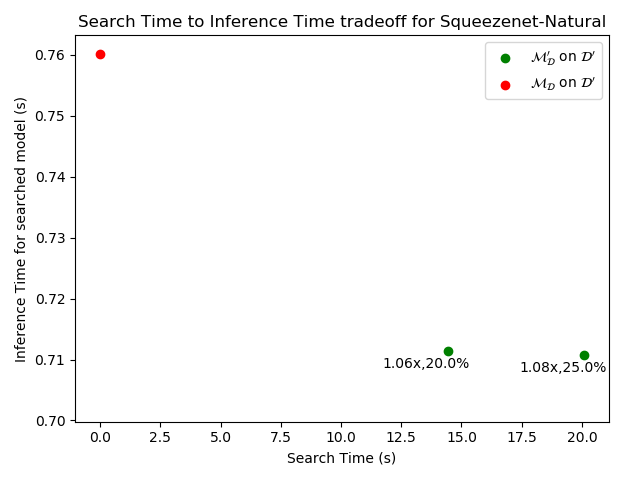} \label{fig:squeezenet_natural_search}} 
    \end{tabular}
    \caption{Trade-off between search time per minibatch and inference time performance of searched model. All models shown here have no accuracy loss compared to $\mathcal{M_D}$ infered on $\mathcal{D'}$. The labels next to the points show (improvement in GOps (-x times), pruning level (\%)) of that model}
    \label{fig:search_inf_tradeoff}
\end{figure*}

%% file: figures/channel_diff.tex
\begin{figure*}[t]
    \centering
    \begin{tabular}{cccc}
        \subfloat[Aquatic - Across networks]{\includegraphics[width = 0.25\textwidth]{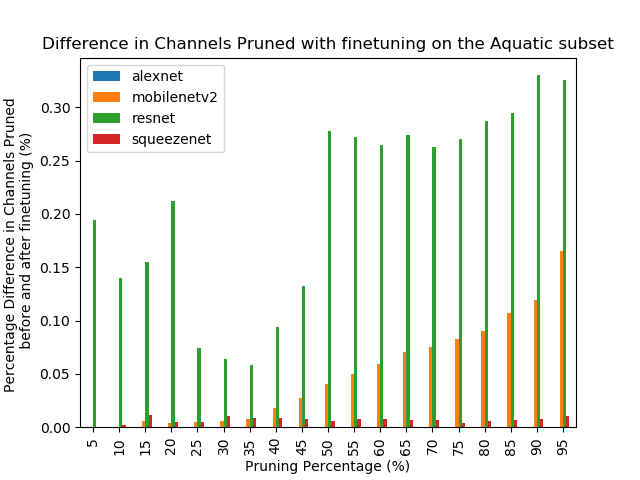} \label{fig:all_nets_aquatic_cdiff}}
        \subfloat[ResNet20 - Aquatic]{\includegraphics[width = 0.25\textwidth]{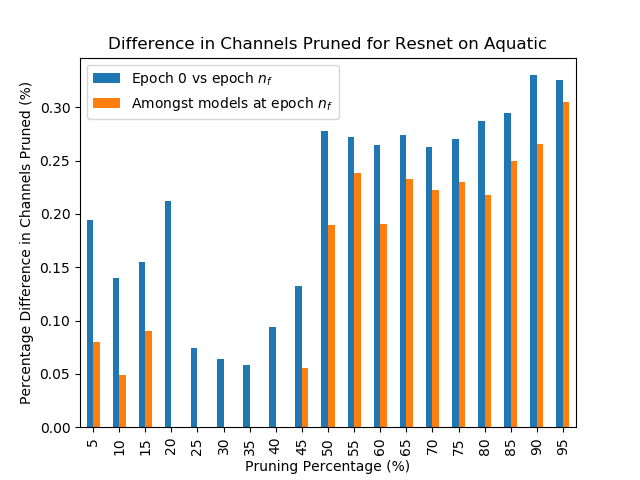} \label{fig:resnet20_aquatic_cdiff}} 
        \subfloat[MobileNetV2 - Indoor]{\includegraphics[width = 0.25\textwidth]{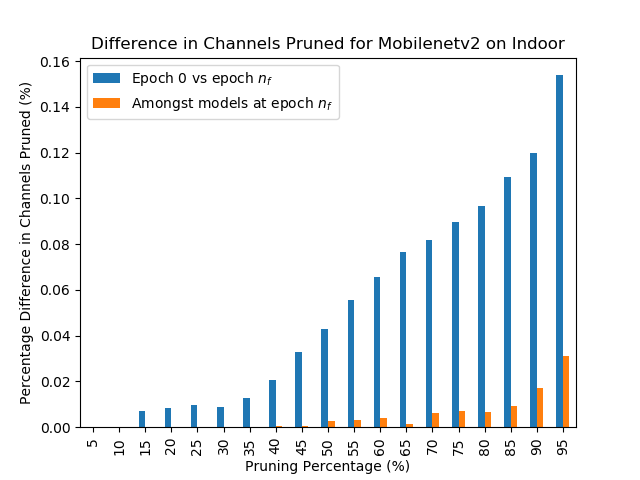} \label{fig:mobilenet_indoor_cdiff}} 
        \subfloat[MobileNetv2 - Random]{\includegraphics[width = 0.25\textwidth]{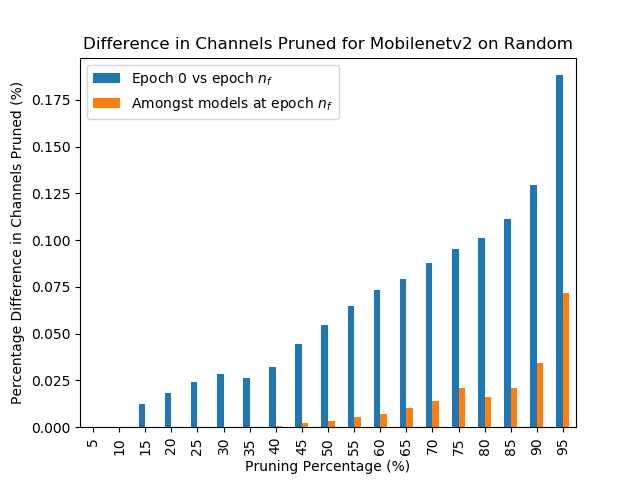} \label{fig:mobilenet_random_cdiff}} 
    \end{tabular}
    \caption{Percentage difference in channels pruned at various points of DaPR for different network and subset combinations}
    \label{fig:cdiff}
\end{figure*}

%% file: main.bbl
\begin{thebibliography}{10}\itemsep=-1pt

\bibitem{imagenet}
{ImageNet}: {A} large-scale hierarchical image database - {IEEE} {Conference}
  {Publication}.

\bibitem{barbu_objectnet_2019}
Andrei Barbu, David Mayo, Julian Alverio, William Luo, Christopher Wang, Dan
  Gutfreund, Josh Tenenbaum, and Boris Katz.
\newblock {ObjectNet}: {A} large-scale bias-controlled dataset for pushing the
  limits of object recognition models.
\newblock page~11.

\bibitem{chollet_xception_2017}
Francois Chollet.
\newblock Xception: {Deep} {Learning} with {Depthwise} {Separable}
  {Convolutions}.
\newblock In {\em 2017 {IEEE} {Conference} on {Computer} {Vision} and {Pattern}
  {Recognition} ({CVPR})}, pages 1800--1807, Honolulu, HI, July 2017. IEEE.

\bibitem{gale_state_2019}
Trevor Gale, Erich Elsen, and Sara Hooker.
\newblock The {State} of {Sparsity} in {Deep} {Neural} {Networks}.
\newblock {\em arXiv:1902.09574 [cs, stat]}, Feb. 2019.
\newblock arXiv: 1902.09574.

\bibitem{gao_fbs_2018}
Xitong Gao, Yiren Zhao, Łukasz Dudziak, Robert Mullins, and Cheng-zhong Xu.
\newblock Dynamic {Channel} {Pruning}: {Feature} {Boosting} and {Suppression}.
\newblock {\em arXiv:1810.05331 [cs]}, Oct. 2018.
\newblock arXiv: 1810.05331.

\bibitem{girshick_fast_2015}
Ross Girshick.
\newblock Fast {R}-{CNN}.
\newblock pages 1440--1448, 2015.

\bibitem{kaiyuan_guo_deephi_2016}
Kaiyuan Guo, Lingzhi Sui, Jiantao Qiu, Song Yao, Song Han, Yu Wang, and
  Huazhong Yang.
\newblock From model to {FPGA}: {Software}-hardware co-design for efficient
  neural network acceleration.
\newblock In {\em 2016 {IEEE} {Hot} {Chips} 28 {Symposium} ({HCS})}, pages
  1--27, Aug. 2016.
\newblock ISSN: null.

\bibitem{resnet_he_deep_2015}
Kaiming He, Xiangyu Zhang, Shaoqing Ren, and Jian Sun.
\newblock Deep {Residual} {Learning} for {Image} {Recognition}.
\newblock {\em arXiv:1512.03385 [cs]}, Dec. 2015.
\newblock arXiv: 1512.03385.

\bibitem{he_resnet_2015}
Kaiming He, Xiangyu Zhang, Shaoqing Ren, and Jian Sun.
\newblock Deep {Residual} {Learning} for {Image} {Recognition}.
\newblock {\em arXiv:1512.03385 [cs]}, Dec. 2015.
\newblock arXiv: 1512.03385.

\bibitem{hua_channel_gated_2019}
Weizhe Hua, Yuan Zhou, Christopher De~Sa, Zhiru Zhang, and G.~Edward Suh.
\newblock Channel {Gating} {Neural} {Networks}.
\newblock {\em arXiv:1805.12549 [cs, stat]}, Oct. 2019.
\newblock arXiv: 1805.12549.

\bibitem{huang_densenet_2018}
Gao Huang, Zhuang Liu, Laurens van~der Maaten, and Kilian~Q. Weinberger.
\newblock Densely {Connected} {Convolutional} {Networks}.
\newblock {\em arXiv:1608.06993 [cs]}, Jan. 2018.
\newblock arXiv: 1608.06993.

\bibitem{hur_entropy-based_2019}
Cheonghwan Hur and Sanggil Kang.
\newblock Entropy-based pruning method for convolutional neural networks.
\newblock {\em The Journal of Supercomputing}, 75(6):2950--2963, June 2019.

\bibitem{iandola_squeezenet_2016}
Forrest~N. Iandola, Song Han, Matthew~W. Moskewicz, Khalid Ashraf, William~J.
  Dally, and Kurt Keutzer.
\newblock {SqueezeNet}: {AlexNet}-level accuracy with 50x fewer parameters and
  {\textless}0.{5MB} model size.
\newblock {\em arXiv:1602.07360 [cs]}, Nov. 2016.
\newblock arXiv: 1602.07360.

\bibitem{jouppi_tpu_nodate}
Norman~P Jouppi, Cliff Young, Nishant Patil, David Patterson, Gaurav Agrawal,
  Raminder Bajwa, Sarah Bates, Suresh Bhatia, Nan Boden, Al Borchers, Rick
  Boyle, Pierre-luc Cantin, Clifford Chao, Chris Clark, Jeremy Coriell, Mike
  Daley, Matt Dau, Jeffrey Dean, Ben Gelb, Tara~Vazir Ghaemmaghami, Rajendra
  Gottipati, William Gulland, Robert Hagmann, C~Richard Ho, Doug Hogberg, John
  Hu, Robert Hundt, Dan Hurt, Julian Ibarz, Aaron Jaffey, Alexander Kaplan,
  Harshit Khaitan, Daniel Killebrew, Andy Koch, Naveen Kumar, Steve Lacy, James
  Laudon, James Law, Diemthu Le, Chris Leary, Zhuyuan Liu, Kyle Lucke, Alan
  Lundin, Gordon MacKean, Adriana Maggiore, Maire Mahony, Kieran Miller, Rahul
  Nagarajan, Ravi Narayanaswami, Narayana Penukonda, Andy Phelps, Jonathan
  Ross, Matt Ross, Amir Salek, Emad Samadiani, Chris Severn, Gregory Sizikov,
  Matthew Snelham, Jed Souter, Dan Steinberg, Andy Swing, Mercedes Tan, Gregory
  Thorson, Bo Tian, Horia Toma, Erick Tuttle, Vijay Vasudevan, Richard Walter,
  Walter Wang, Eric Wilcox, and Doe~Hyun Yoon.
\newblock In-{Datacenter} {Performance} {Analysis} of a {Tensor} {Processing}
  {UnitTM}.
\newblock page~17.

\bibitem{kouris_learning_2018}
Alexandros Kouris and Christos-Savvas Bouganis.
\newblock Learning to {Fly} by {MySelf}: {A} {Self}-{Supervised} {CNN}-{Based}
  {Approach} for {Autonomous} {Navigation}.
\newblock In {\em 2018 {IEEE}/{RSJ} {International} {Conference} on
  {Intelligent} {Robots} and {Systems} ({IROS})}, pages 1--9, Madrid, Oct.
  2018. IEEE.

\bibitem{kouris_informed_2019}
Alexandros Kouris, Christos Kyrkou, and Christos-Savvas Bouganis.
\newblock Informed {Region} {Selection} for {Efficient} {UAV}-based {Object}
  {Detectors}: {Altitude}-aware {Vehicle} {Detection} with {CyCAR} {Dataset}.
\newblock In {\em 2019 {IEEE}/{RSJ} {International} {Conference} on
  {Intelligent} {Robots} and {Systems} ({IROS})}, pages 51--58, Nov. 2019.
\newblock ISSN: 2153-0858.

\bibitem{krizhevsky_alexnet_2017}
Alex Krizhevsky, Ilya Sutskever, and Geoffrey~E. Hinton.
\newblock {ImageNet} classification with deep convolutional neural networks.
\newblock {\em Communications of the ACM}, 60(6):84--90, May 2017.

\bibitem{lecun_mnist_1998}
Y. Lecun, L. Bottou, Y. Bengio, and P. Haffner.
\newblock Gradient-based learning applied to document recognition.
\newblock {\em Proceedings of the IEEE}, 86(11):2278--2324, Nov. 1998.
\newblock Conference Name: Proceedings of the IEEE.

\bibitem{lecun_optimal_1990}
Yann LeCun, John~S. Denker, and Sara~A. Solla.
\newblock Optimal {Brain} {Damage}.
\newblock In D.~S. Touretzky, editor, {\em Advances in {Neural} {Information}
  {Processing} {Systems} 2}, pages 598--605. Morgan-Kaufmann, 1990.

\bibitem{li_l1_pruning_2017}
Hao Li, Asim Kadav, Igor Durdanovic, Hanan Samet, and Hans~Peter Graf.
\newblock Pruning {Filters} for {Efficient} {ConvNets}.
\newblock {\em arXiv:1608.08710 [cs]}, Mar. 2017.
\newblock arXiv: 1608.08710.

\bibitem{lin_runtime_neural_pruning_nodate}
Ji Lin, Yongming Rao, Jiwen Lu, and Jie Zhou.
\newblock Runtime {Neural} {Pruning}.
\newblock page~11.

\bibitem{lin_resnet_pruning_ignore_2019}
Shaohui Lin, Rongrong Ji, Yuchao Li, Cheng Deng, and Xuelong Li.
\newblock Towards {Compact} {ConvNets} via {Structure}-{Sparsity} {Regularized}
  {Filter} {Pruning}.
\newblock {\em arXiv:1901.07827 [cs]}, Mar. 2019.
\newblock arXiv: 1901.07827.

\bibitem{lin_dynamic_2020}
Tao Lin, Luis Barba, Martin Jaggi, Sebastian~U Stich, and Daniil Dmitriev.
\newblock {DYNAMIC} {MODEL} {PRUNING} {WITH} {FEEDBACK}.
\newblock page~22, 2020.

\bibitem{liu_slimming_2017}
Zhuang Liu, Jianguo Li, Zhiqiang Shen, Gao Huang, Shoumeng Yan, and Changshui
  Zhang.
\newblock Learning {Efficient} {Convolutional} {Networks} through {Network}
  {Slimming}.
\newblock Aug. 2017.

\bibitem{segmentation}
Jonathan Long, Evan Shelhamer, and Trevor Darrell.
\newblock Fully {Convolutional} {Networks} for {Semantic} {Segmentation}.
\newblock page~10.

\bibitem{mittal_plasticity_2018}
Deepak Mittal, Shweta Bhardwaj, Mitesh~M. Khapra, and Balaraman Ravindran.
\newblock Recovering from {Random} {Pruning}: {On} the {Plasticity} of {Deep}
  {Convolutional} {Neural} {Networks}.
\newblock {\em arXiv:1801.10447 [cs]}, Jan. 2018.
\newblock arXiv: 1801.10447.

\bibitem{molchanov_sensitvity_importance_nodate}
Pavlo Molchanov, Arun Mallya, Stephen Tyree, Iuri Frosio, and Jan Kautz.
\newblock Importance {Estimation} for {Neural} {Network} {Pruning}.
\newblock page~9.

\bibitem{parashar_scnn_2017}
Angshuman Parashar, Minsoo Rhu, Anurag Mukkara, Antonio Puglielli, Rangharajan
  Venkatesan, Brucek Khailany, Joel Emer, Stephen~W. Keckler, and William~J.
  Dally.
\newblock {SCNN}: {An} {Accelerator} for {Compressed}-sparse {Convolutional}
  {Neural} {Networks}.
\newblock {\em ACM SIGARCH Computer Architecture News}, 45(2):27--40, June
  2017.

\bibitem{sandler_mobilenetv2_2019}
Mark Sandler, Andrew Howard, Menglong Zhu, Andrey Zhmoginov, and Liang-Chieh
  Chen.
\newblock {MobileNetV2}: {Inverted} {Residuals} and {Linear} {Bottlenecks}.
\newblock {\em arXiv:1801.04381 [cs]}, Mar. 2019.
\newblock arXiv: 1801.04381.

\bibitem{simonyan_vgg_2015}
Karen Simonyan and Andrew Zisserman.
\newblock Very {Deep} {Convolutional} {Networks} for {Large}-{Scale} {Image}
  {Recognition}.
\newblock {\em arXiv:1409.1556 [cs]}, Apr. 2015.
\newblock arXiv: 1409.1556.

\bibitem{suau_filter_distillation_2019}
Xavier Suau, Luca Zappella, and Nicholas Apostoloff.
\newblock Filter {Distillation} for {Network} {Compression}.
\newblock {\em arXiv:1807.10585 [cs]}, Dec. 2019.
\newblock arXiv: 1807.10585.

\bibitem{suau_cifar_subset_filter_distill_2019}
Xavier Suau, Luca Zappella, and Nicholas Apostoloff.
\newblock Filter {Distillation} for {Network} {Compression}.
\newblock {\em arXiv:1807.10585 [cs]}, Dec. 2019.
\newblock arXiv: 1807.10585.

\bibitem{szegedy_googlenet_2014}
Christian Szegedy, Wei Liu, Yangqing Jia, Pierre Sermanet, Scott Reed, Dragomir
  Anguelov, Dumitru Erhan, Vincent Vanhoucke, and Andrew Rabinovich.
\newblock Going {Deeper} with {Convolutions}.
\newblock {\em arXiv:1409.4842 [cs]}, Sept. 2014.
\newblock arXiv: 1409.4842.

\bibitem{tan_efficientnet_2019}
Mingxing Tan and Quoc~V. Le.
\newblock {EfficientNet}: {Rethinking} {Model} {Scaling} for {Convolutional}
  {Neural} {Networks}.
\newblock {\em arXiv:1905.11946 [cs, stat]}, Nov. 2019.
\newblock arXiv: 1905.11946.

\bibitem{2020-self-training}
Qizhe Xie, Minh-Thang Luong, Eduard Hovy, and Quoc~V. Le.
\newblock Self-training with {Noisy} {Student} improves {ImageNet}
  classification.
\newblock {\em arXiv:1911.04252 [cs, stat]}, Jan. 2020.
\newblock arXiv: 1911.04252.

\bibitem{xie_resnext_2017}
Saining Xie, Ross Girshick, Piotr Dollár, Zhuowen Tu, and Kaiming He.
\newblock Aggregated {Residual} {Transformations} for {Deep} {Neural}
  {Networks}.
\newblock {\em arXiv:1611.05431 [cs]}, Apr. 2017.
\newblock arXiv: 1611.05431.

\bibitem{zhang_scan_2019}
Linfeng Zhang, Zhanhong Tan, Jiebo Song, Jingwei Chen, Chenglong Bao, and
  Kaisheng Ma.
\newblock {SCAN}: {A} {Scalable} {Neural} {Networks} {Framework} {Towards}
  {Compact} and {Efficient} {Models}.
\newblock {\em arXiv:1906.03951 [cs, stat]}, May 2019.
\newblock arXiv: 1906.03951.

\bibitem{zhang_cambricon-x_2016}
Shijin Zhang, Zidong Du, Lei Zhang, Huiying Lan, Shaoli Liu, Ling Li, Qi Guo,
  Tianshi Chen, and Yunji Chen.
\newblock Cambricon-{X}: {An} accelerator for sparse neural networks.
\newblock In {\em 2016 49th {Annual} {IEEE}/{ACM} {International} {Symposium}
  on {Microarchitecture} ({MICRO})}, pages 1--12, Oct. 2016.
\newblock ISSN: null.

\bibitem{zhao_mayo:_2018}
Yiren Zhao, Xitong Gao, Robert Mullins, and Chengzhong Xu.
\newblock Mayo: {A} {Framework} for {Auto}-generating {Hardware} {Friendly}
  {Deep} {Neural} {Networks}.
\newblock In {\em Proceedings of the 2nd {International} {Workshop} on
  {Embedded} and {Mobile} {Deep} {Learning} - {EMDL}'18}, pages 25--30, Munich,
  Germany, 2018. ACM Press.

\bibitem{zhou_edge_2019}
Zhi Zhou, Xu Chen, En Li, Liekang Zeng, Ke Luo, and Junshan Zhang.
\newblock Edge {Intelligence}: {Paving} the {Last} {Mile} of {Artificial}
  {Intelligence} with {Edge} {Computing}.
\newblock {\em arXiv:1905.10083 [cs]}, May 2019.
\newblock arXiv: 1905.10083.

\bibitem{zmora_distiller_2019}
Neta Zmora, Guy Jacob, Lev Zlotnik, Bar Elharar, and Gal Novik.
\newblock Neural {Network} {Distiller}: {A} {Python} {Package} {For} {DNN}
  {Compression} {Research}.
\newblock {\em arXiv:1910.12232 [cs, stat]}, Oct. 2019.
\newblock arXiv: 1910.12232.

\end{thebibliography}
